\documentclass{article}
\usepackage[inline]{enumitem}
\usepackage{graphicx}
\usepackage{amsmath}
\usepackage{tabularx}
\usepackage{makecell}
\usepackage[dvipsnames]{xcolor}
\usepackage[font=footnotesize,labelfont=bf]{caption}
\usepackage[draft]{hyperref}
\usepackage{natbib}
\usepackage{placeins}
\usepackage[T1]{fontenc}
\usepackage{CJKutf8}
\usepackage{wrapfig,lipsum,booktabs}
\usepackage[english]{babel}

\usepackage[accepted]{icml2017_arxiv}
\usepackage{multirow}
\usepackage{dblfloatfix}

\icmltitlerunning{Sentiment Classification with Word Attention based on Weakly Supervised Learning}

\begin{document}
\begin{CJK}{UTF8}{mj}

\twocolumn[
\icmltitle{Sentiment Classification with Word Attention based on \\ 
	Weakly Supervised Learning with a Convolutional Neural Network}

\begin{icmlauthorlist}
	\icmlauthor{Gichang Lee}{Ko}
	\icmlauthor{Jaeyun Jeong}{Ko}
	\icmlauthor{Seungwan Seo}{Ko}
	\icmlauthor{CzangYeob Kim}{Ko}
	\icmlauthor{Pilsung Kang}{Ko}
\end{icmlauthorlist}

\icmlaffiliation{Ko}{School of Industrial Management Engineering, Korea University, Seoul, South Korea}
\icmlcorrespondingauthor{Pilsung Kang}{pilsung\_kang@korea.ac.kr}

\vskip 0.3in
]

\printAffiliationsAndNotice{}  

\FloatBarrier
\begin{abstract}
In order to maximize the applicability of sentiment analysis results, it is necessary to not only classify the overall sentiment (positive/negative) of a given document but also to identify the main words that contribute to the classification. However, most datasets for sentiment analysis only have the sentiment label for each document or sentence. In other words, there is no information about which words play an important role in sentiment classification. In this paper, we propose a method for identifying key words discriminating positive and negative sentences by using a weakly supervised learning method based on a convolutional neural network (CNN). In our model, each word is represented as a continuous-valued vector and each sentence is represented as a matrix whose rows correspond to the word vector used in the sentence. Then, the CNN model is trained using these sentence matrices as inputs and the sentiment labels as the output. Once the CNN model is trained, we implement the word attention mechanism that identifies high-contributing words to classification results with a class activation map, using the weights from the fully connected layer at the end of the learned CNN model. In order to verify the proposed methodology, we evaluated the classification accuracy and inclusion rate of polarity words using two movie review datasets. Experimental result show that the proposed model can not only correctly classify the sentence polarity but also successfully identify the corresponding words with high polarity scores.

\textbf{Keywords:} \textit{Weakly Supervised Learning, Word Attention, Convolutional Neural Network, Class Activation Mapping}
\end{abstract}
\FloatBarrier

\FloatBarrier
\section{Introduction}
Sentiment analysis and opinion mining is a field of study that analyzes people's opinions, sentiments, evaluations, attitudes, and emotions from written language. It is one of the most active research areas in natural language processing (NLP) and has also been widely studied in data mining, Web mining, and text mining \citep{medhat2014sentiment,liu2012sentiment,pang2008opinion,ravi2015survey} Application domains for sentiment analysis include analyses of customer response to new products or services, analyses of public opinion towards the government’s new policies or political issues under debate, etc. \citep{NODE06607901}. In response to increasing needs in diverse domains, various sentiment analysis techniques have been developed \citep{gui2017learning,cho2014data,poria2016aspect,xianghua2013multi,socher2013recursive,kalchbrenner2014convolutional,tai2015improved}. However, many of the current sentiment analysis techniques suffer from the over-abstraction problem \citep{nasukawa2003sentiment}; the only information obtained from these techniques is the polarity of the document, i.e., whether the nuance of the document is positive or negative. It is difficult to receive more in-depth sentiment analysis results, such as identifying the main words contributing to the polarity classification or finding opposite words or phrase to the overall sentiment of the document, i.e., negative words/phrases in a positive document or positive words/phrases in a negative document.

Recently, attention models have been highlighted in the field of computer vision because of its ability to focus on semantically significant areas in a given image to solve the task of object classification, localization, and detection \citep{ba2014multiple,russakovsky2015imagenet,mnih2014recurrent}. They have also been widely adopted in the field of NLP, as attention models can provide more fruitful interpretations for text analysis tasks \citep{luong2015effective,shen2016attention,rush2015neural}. Attention models help the NLP model focus on salient words/phrases and transfer these attentions to other machine learning models to solve more complicated tasks such as image captioning or text to image generation \citep{xu2015show}. In addition, as one of the basic building blocks of artificial intelligence (AI) is to understand a human speaker’s intention, global technology leaders have released their own AI speakers, such as Amazon’s “Eco,” Google’s “Google Home,” and Apple’s “Homepod,” to collect real-word conversational data in order to upgrade their AI engines. As these AI speakers process the human speaker’s query at a sentence level, it becomes more critical to correctly identify the main intentions (words/phrases) of the speaker, which is the ultimate goal of attention models.

It is not that easy to implement an attention model in NLP tasks. This is mainly because most text datasets have document-level labels, i.e., whether the overall nuance of the document is positive or negative, but phrase- or word-level sentiment labels are rarely available. It implies that there is a restriction that the model should learn attention scores for words or phrases without actual labels. To overcome this problem, previous studies modified the structure of a recurrent neural network (RNN) such that the added weights play an attention role inside the model. Applications of RNN-based attention models include document classification \citep{yang2016hierarchical}, parsing \citep{vinyals2015grammar}, machine translation \citep{bahdanau2014neural,luong2015effective}, and image captioning \citep{xu2015show}.

In this paper, we propose a sentiment classification with a word attention model based on weakly supervised leaning with a convolutional neural network (CNN), named CAM$^2$: Classification and Attention Model with a Class Activation Map. The main advantage of the proposed model is its ability to identify crucial words or phrases in a sentence for the sentiment classification perspective without explicit word- or phrase-level sentiment polarity information. It identifies the words by weak labels only, i.e., the sentence-level polarity that is more abstracted but easily available. In the proposed model, words are embedded in a fixed-size of continuous vector space using Word2Vec \citep{mikolov2013efficient}, GloVe \citep{pennington2014glove}, and FastText \citep{bojanowski2016enriching}. Sentences are represented in a matrix form, whose rows correspond to word vectors, and they are used as the input of a CNN model. The CNN model is trained by considering the sentence-level sentiment polarity as the target, and it produces both the sentence-level polarity score and word-level polarity scores for all words in the sentence, which helps us understand the result of sentence-level sentiment classification. Unlike the existing attention models based on RNN, there is no need to separately learn the weights for the attention. Considering that the same word is used in different contexts for different domains, it is relatively easy to build a dictionary that reflects the characteristics of each domain by using the proposed model.

The rest of this paper is organized as follows. In Section 2, we briefly review and discuss some related works. In Section 3, we demonstrate the architecture of the proposed model. Detailed experimental settings are demonstrated in Section 4 followed by the analysis and discussion of the results. Finally, in Section 5 we present our conclusions.

\section{Related Work}
In this section, we briefly review the representative studies on for CNN-based document classification \citep{kim2014convolutional}, weakly supervised learning for CNN-based object detection \citep{oquab2015object,zhou2016learning}, and the RNN-based document attention model named the hierarchical attention network \citep{yang2016hierarchical}.

\subsection{Convolutional Neural Networks for Document Classification}
\citet{kim2014convolutional} showed CNN, which is the most successful neural network structure for image processing, can also work well for text data, especially for document classification. The architecture of \citet{kim2014convolutional} is shown in Figure 1, and it has the following three main ideas: 

\begin{enumerate}
	\item A large number of filters are used, but the network is not as deep as popular CNN architectures for image processing.
	\item The size of the CNN filter is matched with the vector size of input words.
	\item Multi-channels consisting of static and non-static input vectors are combined. 
\end{enumerate}

Experimental results show that the CNN-based document classification model achieved higher classification accuracies than the conventional machine learning-based models, such as the support vector machine or conditional random field, and other deep neural network structures, such as the deep feedforward neural network or recursive neural network. In addition, the word vector could also be customized for a given corpus, and it sometimes yielded better classification performance than pre-trained word vectors.
\FloatBarrier
\begin{figure*}[t!]
	\centering
	\includegraphics[width=0.7\textwidth]{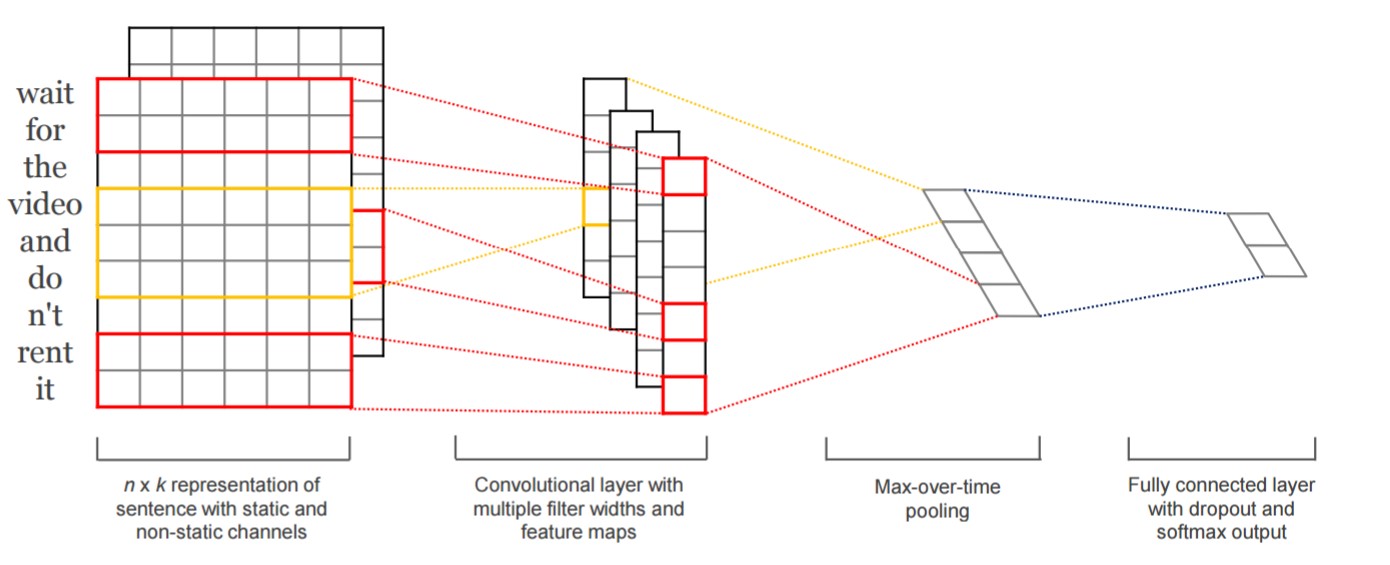}
	\caption{Model architecture with two channels for an example sentence \citep{kim2014convolutional}.}
\end{figure*}

\begin{figure*}[t!]
	\centering
	\includegraphics[width=0.7\textwidth]{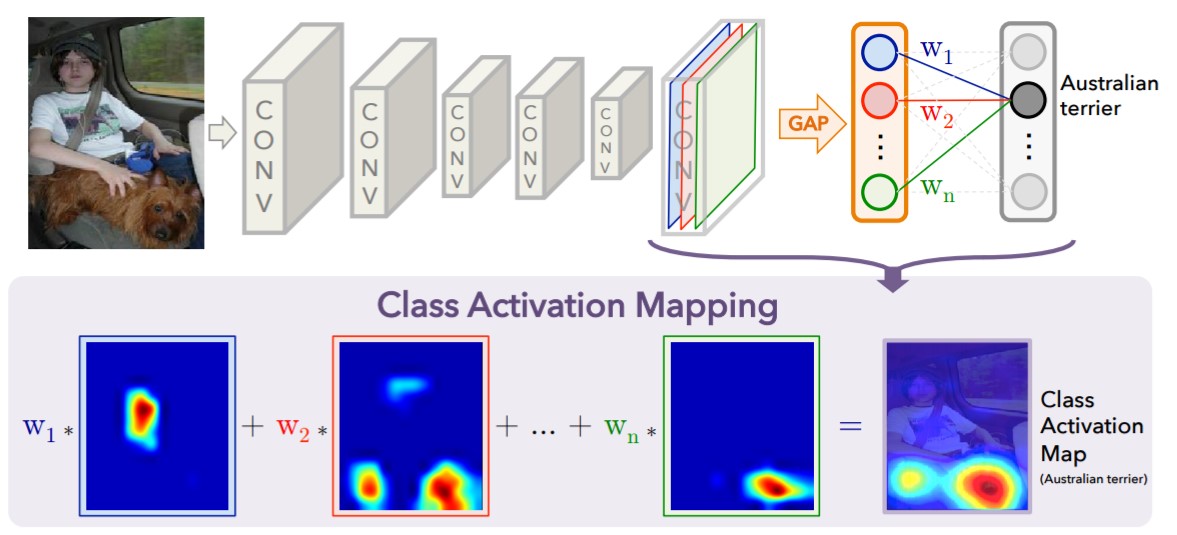}
	\caption{Class activation mapping \citep{zhou2016learning}.}
\end{figure*}

\subsection{Class Activation Mapping}

\begin{figure*}[t!]
	\centering
	\includegraphics[width=1\textwidth]{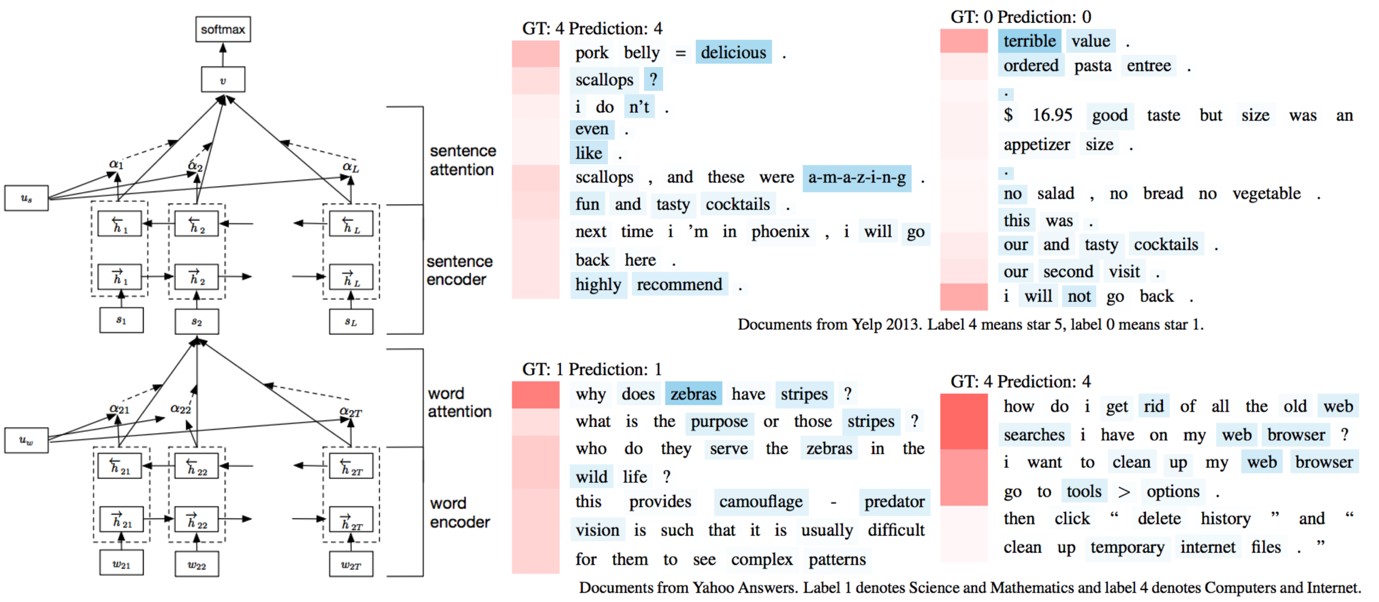}
	\caption{Hierarchical Attention Network \cite{yang2016hierarchical}.}
\end{figure*}

\citet{oquab2015object} proposed a weakly supervised learning method for object detection without bounding box information. In this study, a standard CNN architecture with max pooling between the final convolution and the output layer was utilized. \citet{zhou2016learning} proved the average pooling is more appropriate for the object detection task than the max pooling. The CNN structure and an example of the attention mechanism are shown in Figure 2. In this model, the CNN is trained to correctly classify the object in the input image. In Figure 2, the target of the given image is “Australian terrier,” but no information on the dog’s position in the input image is available during the training. When the training is complete, the weights in the fully connected layers are used to combine the feature map to emphasize the attention area of the original input image. They called this process class activation mapping (CAM), and by utilizing it, not only can the CNN model determine that the “Australian terror” is in the image, but also this classification is mainly inferred by seeing the bottom right part of the image (red area in the final CAM in Figure 2).

\subsection{Hierarchical Attention Network}
\citet{yang2016hierarchical} proposed a hierarchical RNN architecture, inspired by the fact that the document consists of sentences and the sentences are composed of words. In the study, the authors added attention weights to reflect the importance of each sentence and word. As can be seen in Figure 3, the result of their model is the most similar to what we attempted to do in this study. However, the main differences between their work and this work is that \citet{yang2016hierarchical} employed an RNN as the base model and the attention weights were separately learned from the corpus. However, a CNN is employed as the base model for sentiment classification in this study, and we do not explicitly train the model to learn the word-level attention scores.

\section{Classification and Attention Model based on Class Activation Map: CAM$^2$}
\subsection{Overall Framework}
Figure 4 shows the overall framework of the proposed method. After collecting the sentences, low-level embedding is performed by the Word2Vec, GloVe, and FastText methods, and the word vectors in the sentence are concatenated to form the initial input matrix for the CNN. Once the CNN model training is completed, the polarity of a given test sentence is predicted. Then, the weights of the fully connected layer are used to combine the feature maps to produce the attention score for every single word in the sentence.
\begin{figure*}[t!]
	\centering
	\includegraphics[width=\textwidth]{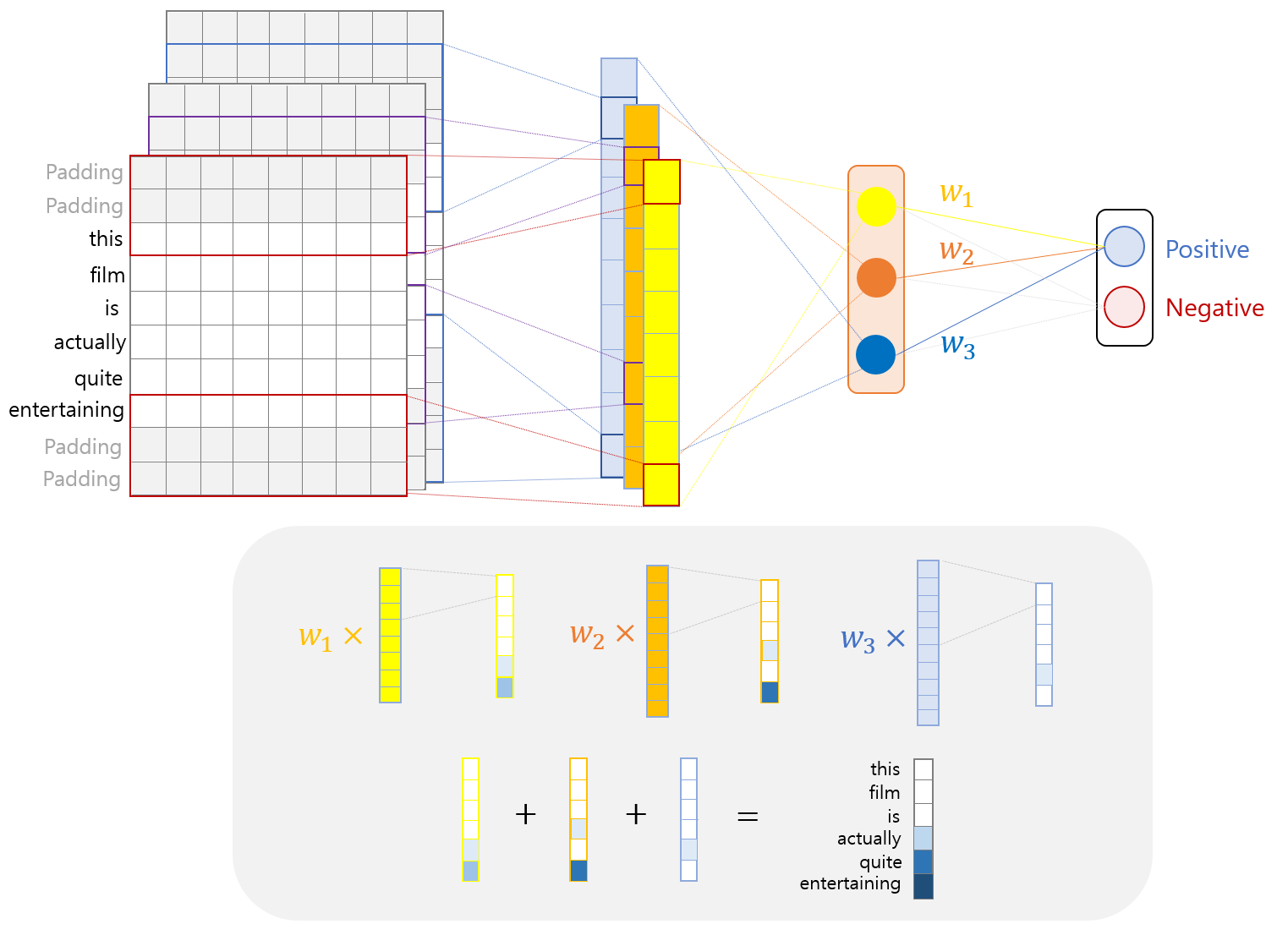}
	\caption{Framework of proposed method.}
\end{figure*}

\subsection{Network Architecture}
The architecture of the CNN used in this paper is basically rooted in the CNN architecture used in \citet{kim2014convolutional}. However, since the CNN used in \citet{kim2014convolutional} was originally designed for document classification, we made some modifications to it to facilitate the extraction of essential words or phrases. First, the zero-padding is added before the first word and after the last word in the sentence to make that the number of times that each word is included in the receptive field during convolution the same, irrespective of the word’s position in the sentence. Second, we applied average-pooling instead of max-pooling. According to \citet{zhou2016learning}, average-pooling and max-pooling are essentially similar, but using average-pooling is advantageous in identifying the overall scope of the target. Third, we increased the number of filters compared to the CAMs used in \citet{oquab2015object} and \citet{zhou2016learning}. As these CAMs are specialized for image processing, the receptive field of convolution is a square (ex: 3 × 3). However, the receptive field of the proposed CAM$^2$ is a rectangular (ex: 3 × word embedding dimension), which integrates a larger amount of information in one scalar value compared to the convolutional filter in image processing. To prevent a possible loss of information due to a larger receptive field, we used a much larger number of convolution filters than was used in \cite{kim2014convolutional}. Finally, we used more various word embedding techniques to form an input matrix of a sentence. \citet{kim2014convolutional} only used the Word2Vec for word embedding, but we consider two recently developed word embedding techniques: GloVe and FastText.

\subsection{Classification and Attention Model based on Class Activation Map}
The input of CNN, $\mathbf{x}_{1:l}$ is created by concatenating the word vectors in a sentence and zero-paddings. We used four type of inputs \textit{CNN-rand, CNN-static, CNN-non-static,} and \textit{CNN-Multichannel}. The \textit{CNN-rand} uses a randomly initialized word vector while \textit{CNN-static} and \textit{CNN-non-static} use the word vectors pre-trained by the Word2Vec. \textit{CNN-Multichannel} uses the word vectors pre-trained by the Word2Vec, GloVe, and FastText. Let $k$, $d$, and $h$ denote the dimension of the word embedding vector, number of maximum words in a sentence, and the height of the receptive field of convolution, respectively, then the input matrix $\mathbf{X} \in R^{([d+2(h-1)]×k)}$ is constructed as follows. The zero-padding is first performed before and after $\mathbf{x}_{1:d}$ so that the number of times that each word is included in the receptive field during convolution is the same ($h$ times).
\begin{eqnarray}
\mathbf{X} = \mathbf{x}_{1:l} = && \underbrace{0 ~ \oplus ~ \mathellipsis ~ \oplus  ~ 0 ~}_\text{$h-1$} \oplus  \nonumber \\ 
&& \underbrace{\mathbf{x}_1 ~ \oplus \mathbf{x}_2 ~ \oplus ~ \mathellipsis ~ \oplus ~ \mathbf{x}_d ~ }_\text{$d$} \oplus \nonumber \\
&& \underbrace{0 ~ \oplus ~ \mathellipsis ~ \oplus ~ 0}_\text{$h-1$}.
\end{eqnarray}
When the window size of the CNN filter, i.e., the height of filter is $h$, the $i$-th feature map $f_i$ is constructed as follows. As the size of CNN filter \textbf{w} is $h \times d$ and zero-padding is performed in the previous step, $f_i$ becomes a $I$-dimensional vector, where $I$ is $(d + h - 1)$.
\begin{eqnarray}
&& \mathbf{f_i}=[f_{1i}, \ f_{2i}, \ \mathellipsis, \ f_{li} ]^T, \\
&& f_{ji}=ReLu(\mathbf{W}_{\mathit{conv}} \odot \mathbf{x}_{j:j+h-1}+b), \\
&& \mathbf{W}_{\mathit{conv}} \in R^{h \times k}, \ b \in R.
\end{eqnarray}
Let $\hat{f_l}$ ̂be the scalar value computed by applying the average pooling to the feature map $\mathbf{f}_i$. The final feature vector $\mathbf{z}$ passed to the fully connected layer is constructed as follows. Considering that n feature maps are computed for a given sentence, $\mathbf{z}$ becomes an $n$-dimensional vector.
\begin{figure*}[t!]
	\centering
	\includegraphics[width=\textwidth]{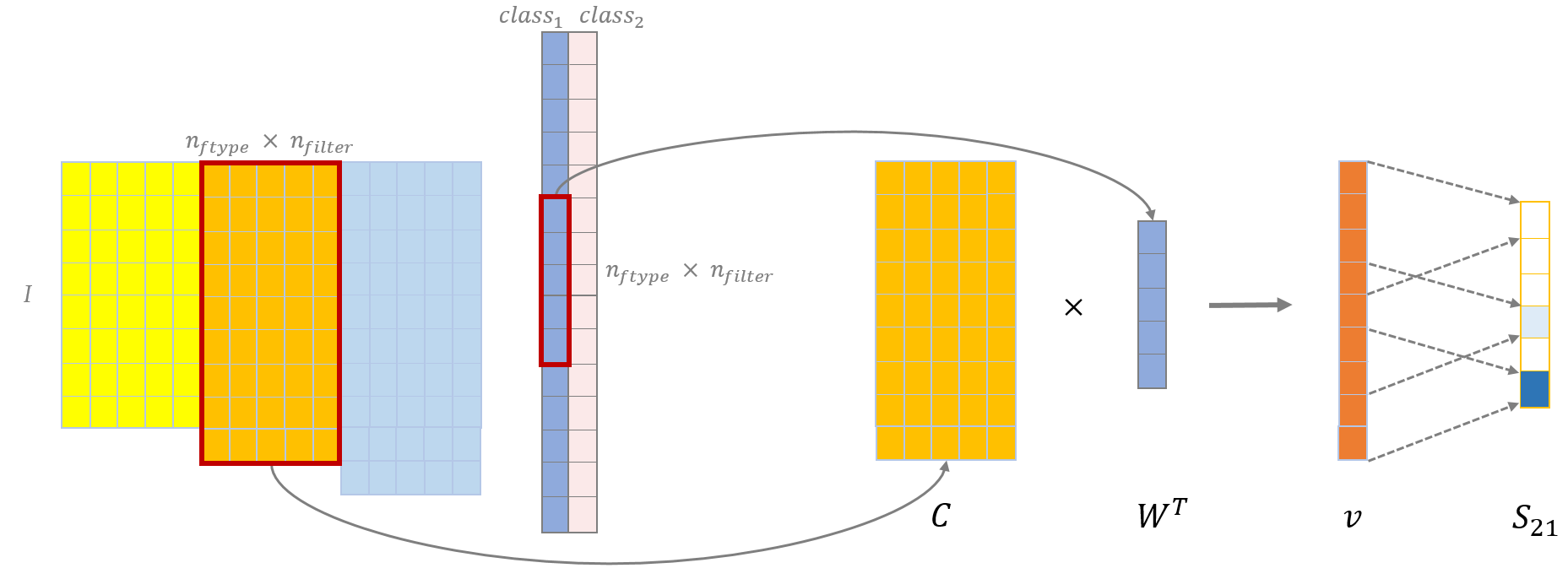}
	\caption{An example of computing a score vector.}
\end{figure*}
\begin{eqnarray}
\mathbf{z}=[\hat{f_1}, \ \hat{f_2}, \ \mathellipsis, \ \hat{f_n}]^T,
\end{eqnarray}
where $n$ is $n_{ftypes}$ (the number of filter type) $\times$ $n_{filters}$ (the number of filters for each type). The output of the fully connected layer for the $i$-th sentence is $\mathbf{y}$, computed as follows:
\begin{eqnarray}
\mathbf{y}=\mathbf{W}_{fc} \cdot \mathbf{z}+\mathbf{b}_{fc}, \\
\mathbf{W}_{fc} \in R^{c\times n}, \\
\mathbf{b}_{fc} \in \ R^c, \\
c: ~ \mathrm{the~number~of~classes.} \nonumber
\end{eqnarray}

Once the CNN model is trained, the sentiment importance score of each word is computed as follows. An illustrated example of the following process is provided in Figure 5. Let $F_l$ be the feature maps corresponding to the $l$-th filter type and $w_{lc_i} $ be the row vector of $\mathbf{W}_{fc}$ for the $l$-th filter type and the $c_i$-th class. Then, the score vector $\mathbf{v}$ is computed as 
\begin{eqnarray}
\mathbf{v}=\mathbf{F}_l \cdot \mathbf{w}_{lc_i}^T, \\
\mathbf{F}_l \in R^{I \times n_{filter}}, \\
\mathbf{w}_{lc_i}^T \in R^{nfilter}.
\end{eqnarray}
The $p$-th element in the score vector $s_{lc_i}$ corresponding to the $l$-th filter type and the $c_i$-th class is computed by averaging $h$ elements with the step size of 1, which makes the $\mathbf{s}_{lc_i}$ a $d$-dimensional vector, regardless of the height of filters:
\begin{eqnarray}
\mathbf{s}_{lc_i} = \frac{1}{h}\sum_{q=p}^{p+h-1}\mathbf{V}_q.
\end{eqnarray}
The final sentiment score of the words in the sentence to $c_i$-th class, $\mathrm{CAM}_{c_i}^2$ is computed by
\begin{eqnarray}
\mathrm{CAM}_{c_i}^2 = \sum_{l=1}^{n_{ftype}}\mathbf{s}_{lc_i}.
\end{eqnarray}

\begin{table*}[t!]
	\begin{center}
		\linespread{1.3}
		\caption{Rating distributions of the IMDB dataset} \label{table1.Rating distributions of the IMDB dataset}%
		\footnotesize{
			\centering{ \setlength\tabcolsep{15pt}
				\begin{tabular}{c|c|c|c|c|c|c|c|c}
					\hline
					Score&  1&  2&  3&  4&   7&   8&   9&   10\\ \hline
					Reviews&  10,122&  4,586&  4,961&  5,531&   4,803&   5,859&   4,607&   9,731\\ \hline
					Class& \multicolumn{4}{c|}{Negative}& \multicolumn{4}{c}{Positive}\\ 
					\hline
		\end{tabular}}}
	\end{center}
\end{table*}

\begin{table*}[t!]
	\begin{center}
		\linespread{1.3}
		\caption{Rating distributions of the WATCHA dataset} \label{table2.Rating distributions of the WATCHA dataset}%
		\footnotesize{
			\centering{ \setlength\tabcolsep{7pt}
				\begin{tabular}{c|c|c|c|c|c|c|c|c|c|c}
					\hline
					score&  0.5&  1&  1.5&  2&   2.5&   3&   3.5&   4& 4.5& 5\\ 
					\hline
					Reviews&  50,660&  66,184&  62,094&  163,272&   173,650&   411,757&   424,378&   652,250& 297,327& 416,096\\ 
					\hline
					Class& \multicolumn{4}{c|}{Negative}&\multicolumn{5}{c|}{Not used}&Positive\\
					\hline
		\end{tabular}}}
	\end{center}
\end{table*}

\begin{table}[t!]
	\begin{center}
		\linespread{1.3}
		\setcounter{table}{3}
		\caption*{The number of tokens} \label{table3. The number of tokens}%
		\footnotesize{
			\centering{ \setlength\tabcolsep{40pt}
				\begin{tabular}{c|c}
					\hline
					IMDB&  WATCHA\\
					\hline
					115,205& 424,027\\
					\hline
		\end{tabular}}}
	\end{center}
\end{table}

\begin{table}[t!]
	\begin{center}
		\linespread{1.3}
		\setcounter{table}{4}
		\caption*{The hyper-parameters of the CNN} \label{table4.The number of tokens}%
		\footnotesize{
			\centering{ \setlength\tabcolsep{18pt}
				\begin{tabular}{c|c}
					\hline
					\multirow{3}{*}{Filter type (window size)} &  3 (tri-gram) \\
					& 4 (quad-gram) \\
					& 5 (5-gram) \\
					\hline
					N. filters & 128 each \\
					\hline
					Doc. length & 100 words \\
					\hline
					Dropout rate & 0.5 \\
					\hline
					$L_2$ regularization ($\lambda$) & 0.1 \\
					\hline
					Batch size & 64 \\
					\hline
		\end{tabular}}}
	\end{center}
\end{table}

\subsection{Word Embedding}
We employed four different word embedding methods to construct the input matrix X: random vectors, Word2Vec, GloVe, and FastText. With the random vectors, the elements of the word vectors were randomly initialized, and they were updated during the CNN training. For the latter three methods, word embedding vectors were separately trained using the same corpus for sentiment classification. We also compared the static word embedding and non-static word embedding methods for CAM$^2$ according to whether the word embedding vectors are updated during the CNN training (non-static) or not (static). In addition, two multi-channel input matrices were also considered. In summary, we tested the following five input matrices for CAM$^2$.
\indent
\begin{enumerate}
\item \textit{CNN-Rand}: word vectors are randomly initialized and they are updated during the CNN training. \label{item:1}
\item \textit{CNN-Static}: word vectors are trained by Word2Vec. They are not updated during the CNN training. \label{item:2}
\item \textit{CNN-Non-Static}: word vectors are trained by Word2Vec first, and they are updated during the CNN training. \label{item:3}
\item \textit{CNN-2ch}: \textit{CNN-Static} and \textit{CNN-Non-Static} are combined. The input of CNN becomes a 3-dimensional (I $\times$ k $\times$ 2) tensor. \label{item:4}
\item \textit{CNN-4ch}: Three matrices with word vectors trained by Word2Vec, GloVe, and FastText are used. They are updated during the CNN training. The \textit{CNN-Non-Static} method is used as the fourth matrix. The input of CNN becomes a 3-dimensional (I $\times$ k $\times$ 4) tensor. \label{item:5}
\end{enumerate}

\section{Experimental Settings}
\subsection{Data Sets \& Target Labeling}
To verify the proposed CAM$^2$, we used two sets of movie reviews, one written in English and the other written in Korean. Not only do movie reviews have explicit sentiment labels (ratings or stars), but they generally also have more subjective expressions compared to other formal texts such as news articles. For the English movie review dataset, we used the publicly available IMDB dataset \citep{maas2011learning}, while Korean movie reviews were collected directly from the WATCHA website (\href{https://watcha.net}{https://watcha.net}), which is the largest movie recommendation service in Korea. Each dataset consists of review sentences and ratings. The distributions of ratings for the IMDB and WATCHA are shown in Table 1 and 2.

\indent
As shown in Table 2, the ratings are well-balanced in the IMDB dataset. Hence, we used the reviews with ratings smaller than or equal to 4 as negative examples, whereas the reviews with ratings greater than or equal to 7 were used as positive examples. Unlike for the IMBD dataset, the ratings of the WATCH dataset are highly skewed toward the positive scores. Therefore, we used the reviews with ratings smaller than or equal to 2 as negative examples whereas only the reviews with 5-point-ratings were used as positive examples. In both datasets, 70\% of the reviews were used as training data, and the remaining 30\% were used as test data.

\subsection{Word Embedding, CNN Parameters, and Performance Measure}

\begin{table}[t!]
	\begin{center}
		\linespread{1.3}
		\setcounter{table}{5}
		\caption*{The test accuracy between methodology} \label{table 5.The test accuracy between methodology}%
		\footnotesize{
			\centering{ \setlength\tabcolsep{15pt}
				\begin{tabular}{c|c|c}
					\hline
					Test & IMDB & WATCHA\\
					\hline
					\textit{CNN-Rand} & 0.8435 & 0.7793 \\
					\textit{CNN-Static} & 0.7750 & 0.7150 \\
					\textit{CNN-Non-Static} & 0.8257 & 0.7538 \\
					\textit{CNN-2channel} & 0.8300 & 0.7602 \\
					\textit{CNN-4channel} & 0.8729 & 0.7533 \\
					\hline
		\end{tabular}}}
	\end{center}
\end{table}

\begin{table}[t!]
	\begin{center}
		\linespread{1.3}
		\setcounter{table}{6}
		\caption*{CAM example} \label{Table 6.CAM example(Normalize it to make the sum of the weights 1).}%
		\footnotesize{
			\centering{ \setlength\tabcolsep{37pt}
				\begin{tabular}{c|c}
					\hline
					Word & Score \\
					\hline
					this & 0.0145 \\
					film & 0.0291 \\
					is & 0.1324 \\
					actually & 0.2183 \\
					quite & 0.2561 \\
					\textbf{entertaining} & \textbf{0.3496} \\
					\hline
		\end{tabular}}}
	\end{center}
\end{table}

\begin{table*}[t!]
	\begin{center}
		\linespread{1.3}
		\caption{Frequently appeared words in the positive/negative sentences in in the IMDB test dataset (semantically positive or negative words are colored in blue and red, respectively)} \label{Table 7.Frequently appeared words in the positive/negative sentences in in the IMDB test dataset (semantically positive or negative words are colored in blue and red, respectively)}%
		\footnotesize{
			\centering{ \setlength\tabcolsep{7pt}
				\begin{tabular}{c c c c c|c c c c c}
					\hline
					\multicolumn{5}{c|}{\textbf{Positive}} & \multicolumn{5}{c}{\textbf{Negative}} \\
					\hline
					\makecell{\textit{CNN-} \\ \textit{Rand}} &
					\makecell{\textit{CNN-} \\ \textit{Static}} &
					\makecell{\textit{CNN-Non-} \\ \textit{Static}} &
					\makecell{\textit{CNN-} \\ \textit{2channel}} &
					\makecell{\textit{CNN-} \\ \textit{4channel}} &
					\makecell{\textit{CNN-} \\ \textit{Rand}} &
					\makecell{\textit{CNN-} \\ \textit{Static}} &
					\makecell{\textit{CNN-Non-} \\ \textit{Static}} &
					\makecell{\textit{CNN-} \\ \textit{2channel}} &
					\makecell{\textit{CNN-} \\ \textit{4channel}} \\
					\hline
					the & and &	and	& and &	and & the & the & the &	the & the \\
					\hline
					and & \color{blue}great & is & is & is & a & is & and & and & and \\
					\hline
					a & is & the & the & a & and & was & \color{BrickRed}worst & \color{BrickRed}worst & of \\
					\hline
					of & a & a & a & the & of & and & of & of & a \\
					\hline
					is & very & of & of & of & to & \color{BrickRed}bad & a & a & \color{BrickRed}worst \\
					\hline
					to & the & s & s & s & is & a & is & is & is \\
					\hline
					I & well & \color{blue}excellent & \color{blue}excellent & it & I & this & the & was & was \\
					\hline
					in & film & it & \color{blue}great & \color{blue}excellent & in & of & \color{BrickRed}awful & \color{BrickRed}awful & I\\
					\hline
					this & of & \color{blue}great & it & to & this & plot & \color{BrickRed}boring & to & to \\
					\hline
					it & it & to & in & I & that & just & was & I & \color{BrickRed}awful\\
					\hline
					that & I & in & to & \color{blue}great & was & acting & to & \color{BrickRed}boring & this\\
					\hline
					was & as & an & I & it & it & movie & I & movie & movie\\
					\hline
					as & \color{blue}excellent & I & an & in & movie & I & \color{BrickRed}bad & this & \color{BrickRed}boring\\
					\hline
					movie & \color{blue}wonderful & was & \color{blue}perfect & \color{blue}perfect & for & \color{BrickRed}awful & \color{BrickRed}poor & \color{BrickRed}poor & \color{BrickRed}bad\\
					\hline
					with & movie & \color{blue}perfect & with & very & with & script & this & \color{BrickRed}bad & s\\
					\hline
					for & story & as & as & was & as & \color{BrickRed}boring & movie & \color{BrickRed}waste & in\\
					\hline
					film & in & \color{blue}best & very & fun & have & to & \color{BrickRed}waste & \color{BrickRed}terrible & \color{BrickRed}waste\\
					\hline
					but & \color{blue}favorite & very & \color{blue}best & by & on & that & \color{BrickRed}terrible & in & \color{BrickRed}poor\\
					\hline
					on & \color{blue}beautiful & \color{blue}enjoyed & \color{blue}enjoyed & \color{blue}enjoyed & film & so & in & s & \color{BrickRed}terrible\\
					\hline
					an & my & with & \color{blue}wonderful & an & but & t & s & with & for\\
					\hline
					have & \color{blue}good & \color{blue}wonderful & \color{blue}fun & as & not & \color{BrickRed}terrible & \color{BrickRed}horrible & are & with\\
					\hline
					are & \color{blue}comedy & \color{blue}fun & by & with & be & it & are & as & as\\
					\hline
					one & \color{blue}loved & by & movie & \color{blue}best & are & \color{BrickRed}stupid & with & by & are\\
					\hline
					his & also & \color{blue}amazing & \color{blue}amazing & \color{blue}wonderful & you & \color{BrickRed}horrible & for & acting & it\\
					\hline
					you & most & movie & \color{blue}loved & \color{blue}loved & an & in & by & for & that\\
					\hline
					not & s & most & that & \color{blue}amazing & at & are & film & \color{BrickRed}horrible & film\\
					\hline
					be & \color{blue}best & \color{blue}loved & \color{blue}superb & most & his & film & as & film & by\\
					\hline
					who & \color{blue}enjoyed & \color{blue}superb & film & are & one & no & it & that & \color{BrickRed}horrible\\
					\hline
					by & \color{blue}love & that & most & for & from & \color{BrickRed}worst & \color{BrickRed}poorly & it & so\\
					\hline
		\end{tabular}}}
	\end{center}
\end{table*}

Each sentence was split into tokens using the space. The punctuations and numbers were removed. All tokens were used to learn the word embedding vectors. We fixed the dimension of word embedding to 100 and set the window size of Word2Vec and FastText to 3. For Word2Vec and FastText, we used the skip-gram structure, while unigram was used to create the co-occurrence matrix for GloVe. The total number of tokens for each dataset is shown in Table 3.

The hyper-parameters for training CNN are summarized in Table 4. We used three different window sizes (how many words are considered in one receptive field), while the number of filters was fixed to 128. The document length, i.e., the maximum number of words, was set to 100. For sentences shorter than 100 words, zero-paddings were added after the last word, whereas the last words were trimmed if sentences were longer than 100 words. We also used two regularization methods. The dropout is an implicit regularization that ignores some weights in each step (dropout rate = 0.5 in this study), whereas the $L_2$ regularization is an explicit regularization that adds the $L_2$-norm of the total weight in the loss function.

\begin{table*}[t!]
	\begin{center}
		\linespread{1}
		\caption{Frequently appearing words in the positive/negative sentences in in the WATCHA test dataset (semantically positive or negative words are in blue and red fonts, respectively)} \label{Table 8. Frequently appearing words in the positive/negative sentences in in the WATCHA test dataset (semantically positive or negative words are in blue and red fonts, respectively)}%
		\footnotesize{
			\centering{ \setlength\tabcolsep{5pt}
				\begin{tabular}{c c c c c|c c c c c}
					\hline
					\multicolumn{5}{c|}{\textbf{Positive}} & \multicolumn{5}{c}{\textbf{Negative}} \\
					\hline
					\makecell{\textit{CNN-} \\ \textit{Rand}} &
					\makecell{\textit{CNN-} \\ \textit{Static}} &
					\makecell{\textit{CNN-Non-} \\ \textit{Static}} &
					\makecell{\textit{CNN-} \\ \textit{2channel}} &
					\makecell{\textit{CNN-} \\ \textit{4channel}} &
					\makecell{\textit{CNN-} \\ \textit{Rand}} &
					\makecell{\textit{CNN-} \\ \textit{Static}} &
					\makecell{\textit{CNN-Non-} \\ \textit{Static}} &
					\makecell{\textit{CNN-} \\ \textit{2channel}} &
					\makecell{\textit{CNN-} \\ \textit{4channel}} \\
					\hline
					
					영화 & 영화 & 영화 & 영화 & 영화 & 영화 & 영화 & 영화 & 영화 & 영화 \\
					\hline
					너무 & 수 & \color{blue}\makecell{최고의 \\ (best)} & \color{blue}\makecell{최고의 \\ (best)} & 너무 & 너무 & 너무 & 너무 & 너무 & 너무\\
					\hline
					이 & \color{blue}\makecell{최고의 \\ (best)} & 너무 & 너무 & \color{blue}\makecell{최고의 \\ (best)} & 이 & 왜 & 이 & 그냥 & 그냥\\
					\hline
					수 & 정말 & 수 & 그리고 & 수 & 더 & \color{BrickRed}\makecell{없고 \\ (none)} & 그냥 & 왜 & 왜\\
					\hline
					왜 & 다시 & 그리고 & 그 & 그 & 왜 & \color{BrickRed}\makecell{없는 \\ (none)} & 왜 & 이 & \color{BrickRed}\makecell{없는 \\ (none)}\\
					\hline
					영화를 & 잘 & 정말 & 수 & 그리고 & 수 & 그냥 & 더 & \color{BrickRed}\makecell{없는 \\ (none)} & 이\\
					\hline
					\color{BrickRed}\makecell{없는 \\	(none)} & 그 & 그 & 가장 & 정말 & \color{BrickRed}\makecell{없다 \\	(none)} & 이 & 그 & 그 & \color{BrickRed}\makecell{없고 \\ (none)} \\
					\hline
					그냥 & 그리고 & 가장 & 정말 & 더 & 그냥 & 좀 & \color{BrickRed}\makecell{없는 \\ (none)} & \color{BrickRed}\makecell{없고 \\ (none)} & 다\\
					\hline
					\color{BrickRed}\makecell{없다 \\ (none)} & 또 & 더 & 더 & 가장 & 영화를 & 영화는 & 영화는 & 더 & 그나마\\
					\hline
					더 & 진짜 & \color{blue}\makecell{최고 \\ (best)} & 이 & 이 & 그 & \color{BrickRed}\makecell{없다 \\ (none)} & 영화를 & 다 & 그\\
					\hline
					다 & 너무 & 있는 & \color{blue}\makecell{최고 \\ (best)} & 잘 & \color{BrickRed}\makecell{없는 \\ (none)} & 이런 & 수 & 정말 & 더\\
					\hline
					이런 & 가장 & 진짜 & 진짜 & 다시 & 다 & 느낌 & \color{BrickRed}\makecell{없고 \\ (none)} & \color{BrickRed}\makecell{없다 \\ (none)} & 정말\\
					\hline
					것 & 내 & 잘 & 다시 & \color{blue}\makecell{최고 \\ (best)} & 것 & \color{BrickRed}\makecell{뻔한 \\ (obvious)} & 좀 & 영화를 & 영화를\\
					\hline
					그 & 있는 & 것 & 잘 & 진짜 & 내가 & 영화를 & 다 & 그나마 & 좀 \\
					\hline
					영화가 & \color{blue}\makecell{좋다 \\ (good)} & 이 & 것 & 것 & 이런 & 보는 & 한 & 좀 & 영화는\\
					\hline
					영화는 & \color{blue}\makecell{아름다운 \\ (beautiful)} & 모든 & 영화가 & 보고 & 영화가 & \color{BrickRed}\makecell{안 \\	(not)} & 정말 & 영화는 & \color{BrickRed}\makecell{없다 \\ (none)} \\
					\hline
					진짜 & 함께 & 영화가 & 보고 & \color{blue}\makecell{좋다 \\ (good)} & 정말 & 내 & \color{BrickRed}\makecell{없다 \\ (none)} & 내 & 이런\\
					\hline
					본 & 더 & 보고 & 있는 & 봐도 & 영화는 & 그 & 그나마 & 수 & 수\\
					\hline
					좀 & 이 & 내 & 내가 & 내 & 이렇게 & 무슨 & 보는 & 이런 & 별\\
					\hline
					정말 & \color{blue}\makecell{최고 \\ (best)} & 다시 & 마지막 & 있는 & 좀 & 건 & 이런 & 대한 & \color{BrickRed}\makecell{별로 \\ (not much of)} \\
					\hline
					내 & 작품 & 내가 & 내 & 본 & 보는 & 것도 & 내 & 보는 & 영화가\\
					\hline
					잘 & 모든 & 봐도 & 모든 & 모든 & 본 & 스토리 & 건 & \color{BrickRed}\makecell{차라리 \\ (rather)} & 내 \\
					\hline
					내가 & 내가 & 본 & 한 & \color{blue}\makecell{완벽한 \\ (perfect)} & 진짜 & 많이 & 대한 & 한 & \color{BrickRed}\makecell{차라리 \\ (rather)} \\
					\hline
					이렇게 & 본 & 내내 & \color{blue}\makecell{좋다 \\ (good)} & 이렇게 & 잘 & \color{BrickRed}\makecell{차라리 \\ (rather)} & \color{BrickRed}\makecell{차라리
						\\ (rather)} & 별 & 안 \\
					\hline
					한 & 봐도 & 또 & \color{blue}\makecell{완벽한 \\ (perfect)} & 내가 & \color{BrickRed}\makecell{없고 \\ (none)} & 것 & 잘 & 건 & \color{BrickRed}\makecell{아깝다 \\ (wasted)} \\
					\hline
					보는 & 중 & \color{blue}\makecell{좋다\\(good)} & 봐도 & 하는 & 한 & \color{BrickRed}\makecell{아닌 \\ (not)} & 별 & 잘 & 느낌 \\
					\hline
					이건 & 있을까 & 마지막 & 영화를 & 다 & 내 & 듯 & 안 & \color{BrickRed}\makecell{뻔한 \\ (obvious)} & 대한\\
					\hline
					\color{blue}\makecell{좋은 \\ (good)} & 꼭 & 대한 & 다 & 또 & 스토리 & 뭘 & 내가 & 봤는데 & \makecell{최악의 \\ (worst)}\\
					\hline
					보고 & 모두 & \color{blue}\makecell{완벽한 \\ (perfect)} & 본 & 한 & 이건 & \color{BrickRed}\makecell{못한 \\ (not)} & 영화가 & \color{BrickRed}\makecell{최악의 \\ (worst)} & 것\\			
					\hline
		\end{tabular}}}
	\end{center}
\end{table*}

\begin{table*}[t!]
	\begin{center}
		\linespread{1.3}
		\caption{Example of word attention for a positively classified sentence in the IMDB dataset} \label{Table 9.Example of word attention for a positively classified sentence in the IMDB dataset}%
		\footnotesize{
			\centering{ \setlength\tabcolsep{1pt}
				\begin{tabular}{m{2.5cm}|l}
					\hline
					\textbf{Methodology} & \multicolumn{1}{c}{\textbf{Sentence}} \\
					\hline
					\multicolumn{1}{r|}{Raw text} & \makecell[l]{I'm normally not a Drama/Feel good movie kind of guy but once I saw the trailer for Radio \\ I couldn't resist. Not only is this a great film but it also has great acting. Cuba Gooding Jr. did \\ an excellent job portraying James Robert Kennedy a.k.a. RAdio. Ed Harris also did a fantastic \\ job as Coach Jones. I was pleasantly surprised to see some comedy in it as well. So for a great \\ story great acting and a little comedy I give Radio a 10 out of 10! (10 / 10 points)} \\
					\hline
					\multicolumn{1}{r|}{\textit{CNN-Rand}} &  \makecell[l]{I m normally not a Drama Feel good movie kind of guy but once I saw the trailer for Radio \\ I couldn t resist Not only is this a great film but it also has great acting Cuba Gooding Jr did \\ an excellent \colorbox{blue}{\textbf{\color{white}job portraying James Robert}} Kennedy a  k a RAdio Ed Harris also did a fantastic \\ job as Coach Jones \colorbox{blue}{\textbf{\color{white}I was pleasantly}} surprised to see some \colorbox{blue}{\textbf{\color{white}comedy in}} it as well So for a great \\ story great acting and a little comedy I give Radio a out of \colorbox{cyan}{\color{white}{Positive}}} \\
					\hline
					\multicolumn{1}{r|}{\textit{CNN-Static}} & \makecell[l]{I m normally not a Drama Feel good movie kind of guy but once I saw the trailer for Radio \\ I couldn t resist Not only is this a \colorbox{blue}{\textbf{\color{white}great film}} but it also has great acting Cuba Gooding Jr did \\ an \colorbox{blue}{\textbf{\color{white}excellent}} job portraying James Robert Kennedy a  k a RAdio Ed Harris also did a \colorbox{blue}{\textbf{\color{white} fantastic}} \\ job as Coach Jones I was pleasantly surprised to see some comedy in it as well So for \colorbox{blue}{\textbf{\color{white}a great}} \\ \colorbox{blue}{\textbf{\color{white} story great acting}} and a little comedy I give Radio a out of \colorbox{cyan}{\color{white}Positive}} \\
					\hline
					\multicolumn{1}{r|}{\textit{CNN-Non-Static}}& \makecell[l]{I m normally not a Drama Feel good movie kind of guy but once I saw the trailer for Radio \\ I couldn t resist Not only is this a great film but it also has great acting Cuba Gooding Jr did \\ \colorbox{blue}{\textbf{\color{white} an excellent job}} portraying James Robert Kennedy a  k a RAdio Ed Harris also did a \colorbox{blue}{\textbf{\color{white} fantastic}}\\ \colorbox{blue}{\textbf{\color{white} job as}} Coach Jones I \colorbox{blue}{\textbf{\color{white} was pleasantly surprised}} to see some comedy in it as well So for a great \\ story great acting and a little comedy I give Radio a out of \colorbox{cyan}{\color{white}Positive}} \\
					\hline
					\multicolumn{1}{r|}{\textit{CNN-2channel}}& \makecell[l]{I m normally not a Drama Feel good movie kind of guy but once I saw the trailer for Radio \\ I couldn t resist Not only is this a great film but it also has great acting Cuba Gooding Jr did \\ \colorbox{blue}{\textbf{\color{white} an excellent job}} portraying James Robert Kennedy a  k a RAdio Ed Harris also did a \colorbox{blue}{\textbf{\color{white} fantastic}} \\ \colorbox{blue}{\textbf{\color{white} job as}} Coach Jones I \colorbox{blue}{\textbf{\color{white} was pleasantly surprised}} to see some comedy in it as well So for a great \\ story great acting and a little comedy I give Radio a out of \colorbox{cyan}{\color{white}Positive}} \\
					\hline
					\multicolumn{1}{r|}{\textit{CNN-4channel}}& \makecell[l]{I m normally not a Drama Feel good movie kind of guy but once I saw the trailer \colorbox{blue}{\textbf{\color{white} for Radio}} \\ I couldn t resist Not only is this a great film but it also has great acting Cuba Gooding Jr did \\ an \colorbox{blue}{\textbf{\color{white} excellent job}} portraying James Robert Kennedy a  k a RAdio Ed Harris also did a \colorbox{blue}{\textbf{\color{white} fantastic}} \\ \colorbox{blue}{\textbf{\color{white} job}} as Coach Jones I \colorbox{blue}{\textbf{\color{white} was pleasantly surprised}} to see some comedy in it as well So for a great \\ story great acting and a little comedy I give Radio a out of \colorbox{cyan}{\color{white}Positive}} \\
					\hline
		\end{tabular}}}
	\end{center}
\end{table*}

\section{Result}
\subsection{Classification Performance}

Table 5 shows the classification accuracies for the five CNN models. It is worth noting that the CNN-Static resulted in the lowest classification accuracy for both IMDB and WATCHA datasets. Since the CNN-Static is the only model which does not update the word embedding vectors during the CNN training, updating the word embedding vectors for a given corpus during the model training, whether or not the word vectors are independently trained before, is encouraged to achieve better classification performance.

Table 6 shows an example of CAM$^2$ for a test sentence. The overall sentiment of this sentence is classified as positive. For each word, the higher the score, the CNN model considers it as a significantly contributing word to the overall sentiment. Thus, the word 'entertaining' had the greatest impact on the classification of this review as being positive.

\subsection{Finding Sentimental Words}

\begin{table*}[t!]
	\begin{center}
		\linespread{1}
		\caption{Example of word attention for a negatively classified sentence in the IMDB dataset} \label{Table 10. Example of word attention for a negatively classified sentence in the IMDB dataset}%
		\footnotesize{
			\centering{ \setlength\tabcolsep{1pt}
				\begin{tabular}{m{2.5cm}|l}
					\hline
					\textbf{Methodology} & \multicolumn{1}{c}{\textbf{Sentence}} \\
					\hline
					\multicolumn{1}{r|}{Raw text} & \makecell[l]{This is one of the most boring films I've ever seen. The three main cast members just didn't \\ seem to click well. Giovanni Ribisi's character was quite annoying. For some reason he seems to\\ like repeating what he says. If he  was the Rain Man it would've been fine but he's not. (3/10 points)} \\
					\hline
					\multicolumn{1}{r|}{\textit{CNN-Rand}} & \makecell[l]{This is one of the most boring films I've ever seen The three main cast members just didn t \\ seem to click well Giovanni Ribisi s character was quite annoying For some reason he seems to like\\ repeating what he says If he was the Rain Man it \colorbox{BrickRed}{\textbf{\color{white}would ve been fine but he s}} not \colorbox{magenta}{\textbf{\color{white}Negative}}} \\
					\hline
					\multicolumn{1}{r|}{\textit{CNN-Static}} & \makecell[l]{This is one of the most \colorbox{BrickRed}{\textbf{\color{white}boring}} films I ve ever seen The three main cast members just didn t\\ seem to click well Giovanni Ribisi s character was \colorbox{BrickRed}{\textbf{\color{white}quite annoying For some reason}} he seems to\\ like repeating what he says If he  was the Rain Man it would ve been fine but he's not \colorbox{magenta}{\textbf{\color{white}Negative}}} \\
					\hline
					\multicolumn{1}{r|}{\textit{CNN-Non-Static}}& \makecell[l]{This is one of the \colorbox{BrickRed}{\textbf{\color{white}most boring films}} I've ever seen The three main cast members just didn t\\seem to click well Giovanni Ribisi s character was quite \colorbox{BrickRed}{\textbf{\color{white}annoying For some}} reason he seems to\\ like repeating what he says If he was the Rain Man it would ve been fine but he's not \colorbox{magenta}{\textbf{\color{white}Negative}}} \\
					\hline
					\multicolumn{1}{r|}{\textit{CNN-2channel}}& \makecell[l]{This is one of the \colorbox{BrickRed}{\textbf{\color{white}most boring films}} I ve ever seen The three main cast members just didn t\\ seem to click well Giovanni Ribisi s character was quite \colorbox{BrickRed}{\textbf{\color{white}annoying For}} some reason he seems to\\ like repeating what he says If he was the Rain Man it would ve been fine but he s not \colorbox{magenta}{\textbf{\color{white}Negative}}} \\
					\hline
					\multicolumn{1}{r|}{\textit{CNN-4channel}}&\makecell[l]{This is one of the \colorbox{BrickRed}{\textbf{\color{white}most boring}} films I ve ever seen The three main cast members just didn t\\seem to click well Giovanni Ribisi s \colorbox{BrickRed}{\textbf{\color{white}character was quite annoying}} For some reason he seems to\\ like repeating what he says If he was the Rain Man it would ve been fine but he s not \colorbox{magenta}{\textbf{\color{white}Negative}}} \\
					\hline
		\end{tabular}}}
	\end{center}
\end{table*}

\begin{table*}[t!]
	\begin{center}
		\linespread{1}
		\caption{Example of word attention for a sentence in the IMDB dataset whose predicted class is different according to CNN models} \label{Table 11.Example of word attention for a sentence in the IMDB dataset whose predicted class is different according to CNN models}%
		\footnotesize{
			\centering{ \setlength\tabcolsep{1pt}
				\begin{tabular}{m{2.5cm}|l}
					\hline
					\textbf{Methodology} & \multicolumn{1}{c}{\textbf{Sentence}} \\
					\hline
					\multicolumn{1}{r|}{Raw text} & \makecell[l]{This movie has a lot to recommend it. The paintings the music and David Hewlett's naked butt are \\ all gorgeous! The plot a story  of redemption forgiveness and courage in the face of adversity is also\\ very interesting and touching -- and it's not predictable which is saying quite a lot about a movie in\\ this day and age. But the acting is mediocre the direction is confusing and the script is just odd. It \\often felt like it was trying to be a parody but I never figured out what it was trying to be parody\\ *of*. (9 / 10 points)} \\
					\hline
					\multicolumn{1}{r|}{\textit{CNN-Rand}} & \makecell[l]{\colorbox{blue}{\textbf{\color{white}This movie has a lot}} to recommend \colorbox{blue}{\textbf{\color{white}it}} The paintings the music \colorbox{BrickRed}{\textbf{\color{white}and David Hewlett s}} naked \\ butt are all gorgeous The plot a story  of redemption \colorbox{blue}{\textbf{\color{white}forgiveness and courage in}} the face of adve-\\rsity is also very interesting and touching and it s not predictable which is saying quite a lot about a \\movie in this day and age But \colorbox{BrickRed}{\textbf{\color{white}the}} acting is mediocre the direction is confusing and the \colorbox{BrickRed}{\textbf{\color{white}script is}}\\ \colorbox{BrickRed}{\textbf{\color{white}just odd It}} often felt like it was trying to be a parody but I never figured out what it was trying \\to be parody of \colorbox{magenta}{\textbf{\color{white}Negative}}} \\
					\hline
					\multicolumn{1}{r|}{\textit{CNN-Static}} & \makecell[l]{This movie has a lot to recommend it \colorbox{blue}{\textbf{\color{white}The paintings the music}} and David Hewlett s naked\\ butt are all gorgeous The plot a story  of redemption forgiveness and courage in the face of adversity\\ \colorbox{blue}{\textbf{\color{white}is also very interesting and touching}} and it s not predictable which is saying quite a lot about\\ a movie in this day and age But the acting \colorbox{BrickRed}{\textbf{\color{white}is mediocre the direction is confusing and the sc-}}\\ \colorbox{BrickRed}{\textbf{\color{white}ript is}} just odd It often felt like it was trying to be a parody but I never figured out what it was \\ trying to be parody of \colorbox{magenta}{\textbf{\color{white}Negative}}} \\
					\hline
					\multicolumn{1}{r|}{\textit{CNN-Non-Static}}& \makecell[l]{This movie has a lot to recommend it \colorbox{blue}{\textbf{\color{white}The paintings}} the \colorbox{blue}{\textbf{\color{white}music and David Hewlett s}} naked \\butt are all gorgeous The plot a story  of redemption forgiveness and courage in the \colorbox{BrickRed}{\textbf{\color{white}face}} of adversity \\is also very interesting \colorbox{blue}{\textbf{\color{white}and touching}} and it s not \colorbox{BrickRed}{\textbf{\color{white}predictable which is}} saying quite a lot about\\ a movie in this day and age But the \colorbox{BrickRed}{\textbf{\color{white}acting is mediocre the direction is}} confusing and the script\\ is just odd It often felt like it was trying to be a parody but I never figured out what it was trying to\\ be parody of \colorbox{cyan}{\color{white}Positive}} \\
					\hline
					\multicolumn{1}{r|}{\textit{CNN-2channel}}& \makecell[l]{This movie has a lot to \colorbox{blue}{\textbf{\color{white}recommend it The paintings}} the music \colorbox{blue}{\textbf{\color{white}and David Hewlett}} s naked\\ butt are all gorgeous The plot a story  of redemption forgiveness and courage in the face of adversity\\ is also very interesting \colorbox{blue}{\textbf{\color{white}and touching and}} it s not \colorbox{BrickRed}{\textbf{\color{white}predictable which is}} saying quite a lot about\\ a movie in this day and age But the \colorbox{BrickRed}{\textbf{\color{white}acting is mediocre the direction is}} \colorbox{BrickRed}{\textbf{\color{white}confusing}} and the sc-\\ ript is just odd It often felt like it was trying to be a parody but I never figured out what it was trying \\to be parody of \colorbox{cyan}{\color{white}Positive}} \\
					\hline
					\multicolumn{1}{r|}{\textit{CNN-4channel}}&\makecell[l]{This movie has a lot to \colorbox{blue}{\textbf{\color{white}recommend it The paintings the music and David Hewlett}} s naked\\ butt are all gorgeous The plot a story of redemption forgiveness and courage in the face of adversity\\is also very interesting and \colorbox{blue}{\textbf{\color{white}touching}} and it s not \colorbox{BrickRed}{\textbf{\color{white}predictable which}} is saying quite a lot about a \\movie in this day and age But the acting \colorbox{BrickRed}{\textbf{\color{white}is mediocre the direction is confusing}} and the \colorbox{BrickRed}{\textbf{\color{white}script}} \\is just odd It often felt like it was trying to be a parody but I never figured out what it was trying to\\ be parody of \colorbox{cyan}{\color{white}Positive}} \\
					\hline
		\end{tabular}}}
	\end{center}
\end{table*}

\begin{table*}[t!]
	\begin{center}
		\linespread{1.3}
		\caption{Example of word attention for a positively classified sentence in the WATCHA dataset} \label{Table 12.Example of word attention for a positively classified sentence in the WATCHA dataset}%
		\footnotesize{
			\centering{ \setlength\tabcolsep{1pt}
				\begin{tabular}{m{2.5cm}|l}
					\hline
					\textbf{Methodology} & \multicolumn{1}{c}{\textbf{Sentence}} \\
					\hline
					\multicolumn{1}{r|}{Raw text} & \makecell[l]{살라딘의 기사도 정신이 진짜 감탄스럽다. 예수상을 다시 세우고 십자가 바닥을 안 밟고 지나가는 장면\\이 존경스럽다. (5 / 5 points)\\
						(Saladin's Chivalry spirit is truly amazing. I’m very impressed by the scene of setting up the Jesus\\ prize and passing without stepping\\ on the floor of the cross.)} \\
					\hline
					\multicolumn{1}{r|}{\textit{CNN-Rand}} & \makecell[l]{살라딘의 기사도 정신이 진짜 감탄스럽다 예수상을 다시 세우고 십자가 바닥을 안 \colorbox{blue}{\textbf{\color{white}밟고}} 지나가는 장\\면이 \colorbox{blue}{\textbf{\color{white}존경스럽다}} \colorbox{cyan}{\color{white}Positive}} \\
					\hline
					\multicolumn{1}{r|}{\textit{CNN-Static}} & \makecell[l]{살라딘의 기사도 정신이 진짜 \colorbox{blue}{\textbf{\color{white}감탄스럽다 예수상을}} 다시 세우고 십자가 바닥을 안 밟고 지나가는\\ 장면이 존경스럽다 \colorbox{cyan}{\color{white}Positive}} \\
					\hline
					\multicolumn{1}{r|}{\textit{CNN-Non-Static}}& \makecell[l]{살라딘의 기사도 정신이 진짜 \colorbox{blue}{\textbf{\color{white}감탄스럽다 예수상을}} 다시 세우고 십자가 바닥을 안 밟고 지나가는\\ 장면이 존경스럽다 \colorbox{cyan}{\color{white}Positive}} \\
					\hline
					\multicolumn{1}{r|}{\textit{CNN-2channel}}& \makecell[l]{살라딘의 기사도 정신이 진짜 \colorbox{blue}{\textbf{\color{white}감탄스럽다 예수상을}} 다시 세우고 십자가 바닥을 안 밟고 지나가는\\ 장면이 존경스럽다 \colorbox{cyan}{\color{white}Positive}} \\
					\hline
					\multicolumn{1}{r|}{\textit{CNN-4channel}}&\makecell[l]{살라딘의 기사도 정신이 진짜 \colorbox{blue}{\textbf{\color{white}감탄스럽다 예수상을}} 다시 세우고 십자가 바닥을 안 밟고 지나가는\\ 장면이 존경스럽다 \colorbox{cyan}{\color{white}Positive}} \\
					\hline
		\end{tabular}}}
	\end{center}
\end{table*}

\begin{table*}[t!]
	\begin{center}
		\linespread{1.3}
		\caption{Example of word attention for a negatively classified sentence in the WATCHA dataset } \label{Table 13.Example of word attention for a negatively classified sentence in the WATCHA dataset }%
		\footnotesize{
			\centering{ \setlength\tabcolsep{1pt}
				\begin{tabular}{m{2.5cm}|l}
					\hline
					\textbf{Methodology} & \multicolumn{1}{c}{\textbf{Sentence}} \\
					\hline
					\multicolumn{1}{r|}{Raw text} & \makecell[l]{영화 전체를 통틀어 가장 불필요하고 의미없는 가오를 잡는 여자가 환호를 받고 있는 아이러니한 영화!\\ 사운드트랙은 인정하더라도 관객을 지나가는 메트로폴리스 행인만도 못하게 다루는 스토리텔링 한마디로 \\총체적 난국.  (2 / 5 points)\\
						(An ironic movie in which the most unnecessary and meaningless flaunt woman in the whole movie is\\ being cheered! Soundtracks are acceptable but storytelling makes the audience run down. A total impa-\\sse in a word.)} \\
					\hline
					\multicolumn{1}{r|}{\textit{CNN-Rand}} & \makecell[l]{영화 전체를 통틀어 가장 \colorbox{BrickRed}{\textbf{\color{white}불필요하고 의미없는 가오를}} 잡는 여자가 환호를 받고 있는 아이러니한\\ 영화 사운드트랙은 인정하더라도 관객을 지나가는 메트로폴리스 행인만도 못하게 다루는 스토리텔링 \\한마디로 총체적 난국 \colorbox{magenta}{\textbf{\color{white}Negative}}} \\
					\hline
					\multicolumn{1}{r|}{\textit{CNN-Static}} & \makecell[l]{영화 전체를 통틀어 가장 불필요하고 의미없는 가오를 잡는 여자가 환호를 받고 있는 아이러니한 영화\\ 사운드트랙은 인정하더라도 관객을 지나가는 메트로폴리스 행인만도 못하게 다루는 스토리텔링\\ \colorbox{BrickRed}{\textbf{\color{white}한마디로 총체적 난국}} \colorbox{magenta}{\textbf{\color{white}Negative}}} \\
					\hline
					\multicolumn{1}{r|}{\textit{CNN-Non-Static}}& \makecell[l]{영화 전체를 통틀어 가장 불필요하고 의미없는 가오를 잡는 여자가 환호를 받고 있는 아이러니한 영화\\ 사운드트랙은 인정하더라도 관객을 지나가는 메트로폴리스 행인만도 못하게 다루는 스토리텔링\\ \colorbox{BrickRed}{\textbf{\color{white}한마디로 총체적 난국}} \colorbox{magenta}{\textbf{\color{white}Negative}}} \\
					\hline
					\multicolumn{1}{r|}{\textit{CNN-2channel}}& \makecell[l]{영화 전체를 통틀어 가장 불필요하고 의미없는 가오를 잡는 여자가 환호를 받고 있는 아이러니한 영화\\ 사운드트랙은 인정하더라도 관객을 지나가는 메트로폴리스 행인만도 못하게 다루는 스토리텔링\\ \colorbox{BrickRed}{\textbf{\color{white}한마디로 총체적 난국}} \colorbox{magenta}{\textbf{\color{white}Negative}}} \\
					\hline
					\multicolumn{1}{r|}{\textit{CNN-4channel}}&\makecell[l]{영화 전체를 통틀어 가장 불필요하고 의미없는 가오를 잡는 여자가 환호를 받고 있는 아이러니한 영화\\ 사운드트랙은 인정하더라도 관객을 지나가는 메트로폴리스 행인만도 못하게 다루는 스토리텔링\\ \colorbox{BrickRed}{\textbf{\color{white}한마디로 총체적 난국}} \colorbox{magenta}{\textbf{\color{white}Negative}}} \\
					\hline
		\end{tabular}}}
	\end{center}
\end{table*}

\begin{table*}
	\begin{center}
		\linespread{1.3}
		\caption{Example of word attention for a sentence in the IMDB dataset whose predicted class is different according to CNN models} \label{Table 14.Example of word attention for a sentence in the IMDB dataset whose predicted class is different according to CNN models}%
		\footnotesize{
			\centering{ \setlength\tabcolsep{1pt}
				\begin{tabular}{m{2.5cm}|l}
					\hline
					\textbf{Methodology} & \multicolumn{1}{c}{\textbf{Sentence}} \\
					\hline
					\multicolumn{1}{r|}{Raw text} & \makecell[l]{이렇게 재미없고 그래픽도 꾸지고 난장판인 엑스맨을 과거의 이야기로 새로 시작한 메튜 본 감독과\\ 깔끔하게 다시 재정리한 브라이언 싱어 감독에게 박수를… ( 1 / 5 points)\\
						(I would like to pay tribute to Bryan Singer, who just reconstituted this boring and messy X-Men as\\ a story of the past, and Matthew Vaughn, who neatly rearranged it again.)
					}\\
					\hline
					\multicolumn{1}{r|}{\textit{CNN-Rand}} & \makecell[l]{이렇게 재미없고 그래픽도 꾸지고 난장판인 엑스맨을 과거의 \colorbox{OliveGreen}{\textbf{\color{white}이야기로}} 새로 \colorbox{OliveGreen}{\textbf{\color{white}시작한}} 메튜 본 \\감독과 깔끔하게 다시 재정리한 브라이언 \colorbox{BrickRed}{\textbf{\color{white}싱어}} 감독에게 \colorbox{BrickRed}{\textbf{\color{white}박수를}} \colorbox{magenta}{\textbf{\color{white}Negative}}} \\
					\hline
					\multicolumn{1}{r|}{\textit{CNN-Static}} & \makecell[l]{이렇게 \colorbox{BrickRed}{\textbf{\color{white}재미없고 그래픽도}} 꾸지고 난장판인 엑스맨을 과거의 이야기로 새로 시작한 메튜 본 \\감독과 깔끔하게 다시 재정리한 브라이언 싱어 \colorbox{OliveGreen}{\textbf{\color{white}감독에게}}  \colorbox{OliveGreen}{\textbf{\color{white}박수를}} \colorbox{cyan}{\color{white}Positive}} \\
					\hline
					\multicolumn{1}{r|}{\textit{CNN-Non-Static}}& \makecell[l]{이렇게 \colorbox{BrickRed}{\textbf{\color{white}재미없고 그래픽도}} 꾸지고 난장판인 엑스맨을 과거의 이야기로 새로 시작한 메튜 본\\ 감독과 \colorbox{OliveGreen}{\textbf{\color{white}깔끔하게 다시}} 재정리한 브라이언 싱어 감독에게 박수를 \colorbox{magenta}{\textbf{\color{white}Negative}}} \\
					\hline
					\multicolumn{1}{r|}{\textit{CNN-2channel}}& \makecell[l]{이렇게 \colorbox{BrickRed}{\textbf{\color{white}재미없고 그래픽도 꾸지고}} 난장판인 엑스맨을 과거의 이야기로 새로 시작한 메튜 본 \\감독과 \colorbox{OliveGreen}{\textbf{\color{white}깔끔하게 다시}} 재정리한 브라이언 싱어 감독에게 박수를 \colorbox{magenta}{\textbf{\color{white}Negative}}} \\
					\hline
					\multicolumn{1}{r|}{\textit{CNN-4channel}}&\makecell[l]{이렇게 \colorbox{BrickRed}{\textbf{\color{white}재미없고 그래픽도 꾸지고}} 난장판인 엑스맨을 과거의 이야기로 새로 시작한 메튜 본 \\감독과 \colorbox{OliveGreen}{\textbf{\color{white}깔끔하게 다시}} 재정리한 브라이언 싱어 감독에게 박수를 \colorbox{magenta}{\textbf{\color{white}Negative}}} \\
					\hline
		\end{tabular}}}
	\end{center}
\end{table*}

Table 7 provides the frequent words listed in the IMDB test dataset by selecting the top five highly scored words in the sentences classified as positive (left five columns) and negative (right five columns). It is worth noting that although the CNN-Rand yielded a relatively good classification performance compared to other techniques, it identified the least emotional words among the five CNN models. Although the classification performance of CNN-Static was the worst, its attention mechanism seemed to work well, in that many emotional words were highly ranked. In terms of classification performance, it is important whether or not the input vector is updated in the training process. However, for the sake of word attention in sentiment classification, it becomes more important whether the general grammatical relationship between the words are well-preserved in the word embedding vector (not updated for classification task).

Table 8 provides the frequent words listed in the WATCHA test dataset by selecting the top five highly scored words in the sentences classified as positive (left five columns) and negative (right five columns). In this case, the emotional word in the upper word list is somewhat overlapped with other methods compared to the IMDB dataset. This is because Korean is an agglutinative language, which tends to have a high rate of affixes per word. For example, “없다, 없는, 없고…(none),” “안, 아닌, 못…(not),” and “차라리(rather)” are usually used in Korean for negative expressions. Experimental results confirm that these words are more frequently used in the negative reviews than in the positive reviews (except CNN-Rand).

\subsection{Word Attention: IMDB}
Table 9 shows an example of word attention of a positively classified sentence in the IMDB dataset. The words highlighted in blue are the top 10\% highly scored words in the sentence. The four models except the \textit{CNN-Rand} can successfully capture semantically positive words or phrases (ex. excellent, fantastic, and was pleasantly surprised). In particular, the \textit{CNN-Static} is especially good at paying attention to longer sentimental phrases such as “a great story great acting.”

Table 10 shows an example of word attention of a negatively classified sentence in the IMDB dataset. The words highlighted in red are the top 10\% highly scored words in the sentence. If one reads the review, he/she can easily recognize multiple negative expressions within the review, which results in different attention words or phrases according to different models. For example, the \textit{CNN-Non-Static, CNN-2channel,} and \textit{CNN-4channel} pay attention to “boring” and “annoying,” both of which are clearly negative expressions when used in a movie review. However, there is another explicit negative expression, namely, “it would (have) been fine,” which receives an attention by the \textit{CNN-Rand}.

Table 11 shows an example of attention results for a sentence whose predicted class is different according to the CNN models because of mixed emotional expressions within the sentence. In this case, the words in the top 10\% highest scores are highlighted in blue and those in the bottom 10\% lowest scores are highlighted in red if the sentence is classified as positive. The highlighting scheme is reversed if the sentence is classified as negative. Likewise, the \textit{CNN-Static, CNN-Non-Static, CNN-2channel,} and \textit{CNN-4channel} have relatively better attention performances than the CNN-Rand. Again, the \textit{CNN-Static} has a relatively good performance in capturing longer emotional phrases such as “is also very interesting and touching.”

\subsection{Word Attention: WATCHA}
Table 12 shows an example of word attention of a positively classified sentence in the WATCHA dataset. The words highlighted in blue are the top 10\% highly scored words in the sentence. In this sentence, there are two obvious positive expressions, i.e., 감탄스럽다 (impressing) and 존경스럽다 (admirable); the former was successfully detected by \textit{CNN-Static, CNN-Non-Static, CNN-2channel,} and \textit{CNN-4channel} while the latter was detected by \textit{CNN-Rand}.

Table 13 shows an example of word attention of a negatively classified sentence in the WATCHA dataset. The words highlighted in blue are the top 10\% highly scored words in the sentence. This sentence also has two semantically explicit negative expressions: “불필요하고 의미없는 가오 (unnecessary and meaningless flaunt)” and “한마디로 총체적 난국 (a total crisis in a word).” The \textit{CNN-Rand} focused on the former expression, whereas the rest of the four models focused on the latter expression. Similar to the example of the positive sentence in Table 12, it seems that the attention mechanism of \textit{CNN-Rand} is somewhat different from those of the other models. This is mainly because the word embedding vectors are not updated to reflect the user’s rating information. Hence, more general emotional expressions, rather than movie-review specific expressions, receive higher attention by the \textit{CNN-Rand}.

Table 14 shows an examples in the same manner as the example illustrated in Table 11. The three models except \textit{CNN-Rand} and \textit{CNN-Static} focus on the negative phrase “재미없고 (boring)” and the positive phrase “깔끔하게 (neatly)”. Qualitatively, the former is a stronger emotional expression than the latter, which results in the entire sentence being predicted as negative. However, the CNN-Static finds a stronger positive expression, i.e., “박수를 (pay tribute to)” rather than “깔끔하게 (neatly)”, which results in the CNN model predicting the whole sentence as positive.

\section{Conclusion}
In this paper, we propose $CAM^2$, a classification and attention model with class activation map, which is a sentiment classification model with word attention based on weakly supervised CNN learning. Although the proposed model is trained based on class labels only, it can not only predict the overall sentiment of a given sentence but also find important emotional words significantly contributing the predicted class. Compared to the previous CNN-based text classification model, $CAM^2$ utilizes zero-paddings to help the CNN consider every word equally regardless of its position in the sentence. Moreover, it uses average pooling and a large number of filters to preserve the information as much as possible. In addition, various word embedding techniques are employed and integrated.\\
Experimental results on two movie review datasets, IMDB, which is in English, and WATCHA, which is in Korean, show that the proposed $CAM^2$ yielded classification accuracies higher than 87\% for the IMDB and 78\% for the WATCHA dataset. The CNN models that update the word embedding vectors during the sentiment classification learning (\textit{CNN-Rand, CNN-Non-Static, CNN-2channel,} and \textit{CNN-4channel}) achieved higher classification performance than that did not update the word embedding vectors (\textit{CNN-Static}). It is also worth noting that the integration of  multiple word embedding techniques improved the classification performance for the IMDB dataset. However, all models showed the ability to find important emotional words in the sentence, although the internal mechanism might be  different. For the WATCHA dataset, in particular, the \textit{CNN-Static}, which does not update the word embedding vector during the training, focused more on generally accepted emotional expressions, whereas the other models, which adapt to the language usage pattern in the movie review domain, seemed to focus more on the domain-dependent emotional expressions.\\
We expect that the proposed methodology can be a useful application in domains where it is important to understand what the input sentences are intended to convey, such as visual question and answering system or chat bots. Although the experimental results were favorable, the current study has some limitations, which lead us to the future research directions. First, the proposed method used a simple space-based token for training word embedding vectors. If more sophisticated preprocessing techniques, such as lemmatization, are performed, the classification and attention performance can be improved. Secondly, quantitative evaluation of word attention, i.e., how good or appropriate the identified words are in the context of sentiment classification, is difficult, which is why we qualitatively interpreted the word attention results in Section 4. Developing a systematic and quantitative evaluation method for word attention can be another meaningful future research topic.\\

\bibliographystyle{icml2017}
\bibliography{References}
\end{CJK}
\end{document}